\documentclass[pdflatex,sn-mathphys-num]{sn-jnl}

\usepackage{graphicx}%
\usepackage{multirow}%
\usepackage{amsmath,amssymb,amsfonts}%
\usepackage{amsthm}%
\usepackage{mathrsfs}%
\usepackage[title]{appendix}%
\usepackage{xcolor}%
\usepackage{textcomp}%
\usepackage{manyfoot}%
\usepackage{booktabs}%
\usepackage{algorithm}%
\usepackage{algorithmicx}%
\usepackage{algpseudocode}%
\usepackage{listings}%
\usepackage{caption}%
\usepackage{subcaption}%

\theoremstyle{thmstyleone}%
\theoremstyle{thmstyletwo}%

\theoremstyle{thmstylethree}%

\begin{document}

\title[Article Title]{Topic-Aware Multi-Task Mental Health Classification with 
Active Learning-Based Stepwise Relabeling for Social Media Conversations}


\author*[1]{\href{https://orcid.org/0009-0008-9101-8076}{\fnm{Revathy Iswarya} \sur{Sekaran}}}\email{riswarya@umich.edu}

\author[2]{\href{https://orcid.org/0009-0005-7824-9302}{\fnm{Vansh} \sur{Bhardwaj}}}\email{vanshbhardwaj1018@gmail.com}

\author[3]{\href{https://orcid.org/0009-0001-0167-8564}{\fnm{Aveek} \sur{Sahana}}}\email{sahana.aveek@gmail.com}

\author[1]{\href{https://orcid.org/0000-0002-1168-0684}{\fnm{Utkarshani} \sur{Jaimini}}}\email{jaimini@umich.edu}

\affil*[1]{\orgname{University of Michigan-Dearborn}, \orgaddress{\city{Dearborn}, \state{Michigan}, \country{USA}}}

\affil[2]{\orgname{JK Lakshmipat University}, \orgaddress{\city{Jaipur}, \state{Rajasthan}, \country{India}}}

\affil[3]{\orgaddress{\city{Bentonville}, \state{Arkansas}, \country{USA}}}


\abstract{
Mental health conditions affect millions of people worldwide, yet many cases go undetected until they reach a critical stage. Social media platforms like Reddit have become spaces where people openly share their experiences, but the support is often provided by individuals without clinical training. Existing multi-task approaches jointly model emotion and sentiment for mental health classification but frequently rely on noisy labels generated by pre-trained models and overlook topic information as a useful signal. This paper introduces two key contributions: First, an existing Core Fusion Network framework is extended by adding Topic alongside existing Mental Health Condition, Emotion, and Sentiment Classification tasks, with topic labels derived from the therapist-client interactions in the CounselChat dataset. Second, an Active Learning-based Stepwise Relabeling approach is proposed to refine a small subset of noisy emotion and topic labels using pre-trained models in two stages: targeting medium-uncertainty samples first, followed by high-uncertainty samples. The predictions above a confidence threshold (0.7) are accepted to prevent error propagation. The experiments are conducted on the SuicideWatch and Mental Health dataset of 54,412 Reddit posts collected from five mental health subreddits, evaluating domain-specific transformer encoders including MentalBERT and MentalRoBERTa. The proposed Active Learning-based Stepwise Relabeling, when applied to Core Fusion Network with MentalRoBERTa, improves Mental Health Condition by 0.09\%, Emotion by 0.51\%, Sentiment by 0.35\%, and Topic Classification by 1.57\% over the baseline Core Fusion Network. The results suggest that the proposed framework significantly reduces label noise in multi-task mental health classification, and that topic information provides additional context.
}




\maketitle

\section{Introduction}\label{sec1}
Mental health is defined as a person's ability to deal with everyday challenges, make the most of their abilities, learn and work effectively, and contribute to their community~\cite{bib1}, yet related issues remain a major global health burden. Mental disorders now account for one in six years lived with disability globally, with productivity losses far exceeding direct care costs~\cite{bib2}. Yet, only one-third of people with depression receive formal care due to stigma, discrimination, and under-resourced health systems~\cite{bib3}. As of 2023, nearly 1.2 billion people worldwide were living with a mental disorder~\cite{bib4}. Among those aged 10--19, one in seven experiences a mental disorder, with suicide remaining the third leading cause of death among those aged 15--29~\cite{bib5}.

In recent years, social media platforms have changed how people discuss and seek help for mental health issues, with studies continually showing links between social media use and higher rates of depression, anxiety, and psychological distress, particularly among adolescents \cite{bib6}. Reddit stands out as a platform where people openly discuss their mental health and seek support from peers, creating a large collection of authentic mental health conversations. The peer supporters on these platforms are rarely clinicians, so there is a clear need for reliable NLP models that can accurately classify mental health conditions from social media text \cite{pathak2025}.

Domain-specific transformer models like MentalBERT \cite{ji2021mentalbertpubliclyavailablepretrained} have shown strong performance in detecting mental health-related patterns in social media text. Multi-task learning has further improved results by jointly training related affective tasks alongside mental health condition identification \cite{ruder2017overviewmultitasklearningdeep}. The Core Fusion Network (CFN), introduced by Pathak et al. \cite{pathak2025}, adopts a multi-task methodology on the MotiVAte dataset by combining mental health disorder identification with emotion and sentiment analysis tasks. However, its auxiliary task labels are generated automatically by pre-trained models and are not updated during training, so any errors in these labels can carry through the shared layers and affect performance. The framework also does not explore topic as a useful auxiliary task.

To address these limitations, this work presents an enhanced multi-task learning framework for social media-based mental health classification. By integrating clinician-guided topic information alongside emotional and sentiment cues, the extended Core Fusion Network captures richer context from user posts. In addition, an Active Learning-based Stepwise Relabeling approach is introduced to mitigate label noise while preserving previously learned task representations.

\begin{figure}[H]
\centering
\includegraphics[width = \textwidth,]{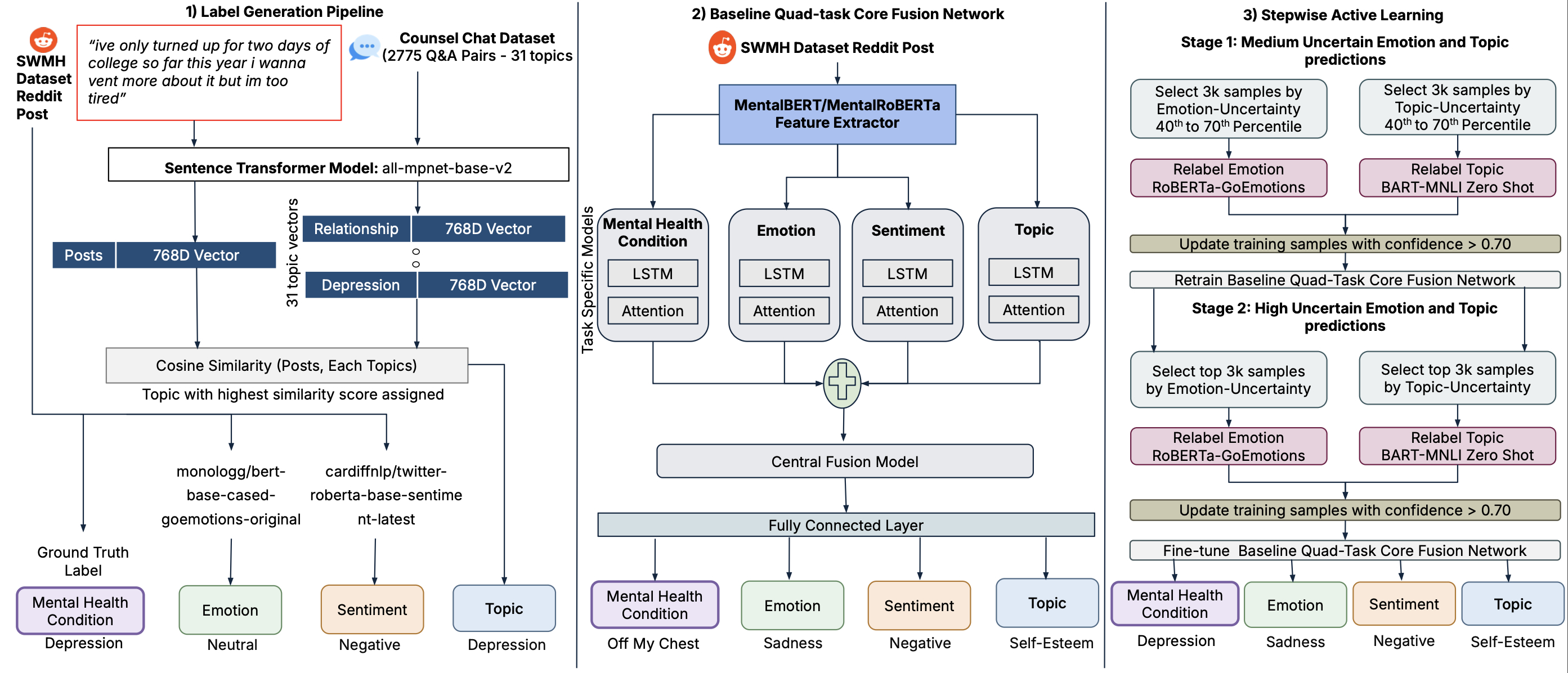}
\caption{\textbf{Overview of the proposed framework. (1) Label Generation using pre-trained models for emotion, sentiment and topic labels. (2) The extended Core Fusion Network for multi-task prediction. (3) The Active Learning-based Stepwise Relabeling approach.}}
\label{fig:pipeline_overview}
\end{figure}

The main contributions of this work are as follows:

\begin{enumerate}
\renewcommand{\labelenumi}{(\alph{enumi})}

\item The existing Core Fusion Network (CFN) is extended by adding Topic Classification alongside existing Mental Health Condition, Emotion, and Sentiment Classification tasks. The emotion and sentiment labels are generated by pre-trained models, while topic labels are derived from clinically meaningful therapist-client interactions in the CounselChat \cite{bertagnolli2020counsel} dataset.
\item The Active Learning-based Stepwise Relabeling approach is proposed to refine a small subset of noisy emotion and topic labels using pre-trained models in two stages: targeting medium-uncertainty samples first, followed by high-uncertainty samples. The baseline Core Fusion Network is retrained from scratch on the updated training set in Stage 1 and fine-tuned in Stage 2.
\item Extensive experiments are conducted on the SuicideWatch and Mental Health \cite{swmh} dataset of 54,412 Reddit posts, including an ablation study of Active Learning strategies and comparisons against multi-task baselines and LLM prompting.
\item Domain-specific transformer encoders, including MentalBERT and MentalRoBERTa, are evaluated as feature extractors to capture clinically relevant linguistic representations for multi-task mental health classification.
\end{enumerate}

\section{Related Works}
The recent advances in the Natural Language Processing(NLP) have significantly improved medical health analysis, especially with the mental health related tasks. The following are some of the important works that have been done in the field of mental health:

\par \textit{Pathak et al.} proposed the \textbf{Core Fusion Network (CFN)} \cite{pathak2025}, a multi task learning framework for Mental Health Disorder Identification (MHDI) using conversational data. Their framework jointly performs three tasks: Mental Health Disorder Identification, Emotion Recognition, and Sentiment Analysis. It uses separate task specific models along with a central fusion network(CFN) to learn both shared and task specific features. To support this framework, the authors extended the \textbf{MotiVAte} dataset by automatically generating emotion and sentiment labels using pre-trained language models. They have also introduced the CFN architecture, which combines information from all three tasks to improve the overall performance. However, the framework is limited to three tasks and does not consider the topic of the conversation, which can provide additional context for understanding a person's mental health condition.

\par Domain specific transformers have also played an important role in improving mental health analysis. While models such as BERT and RoBERTa have achieved strong performance on many Natural Language Processing tasks, they are trained on general text and they don't fully capture the language and context of mental-health talks. To improve upon this, researchers introduced \textbf{MentalBERT} and \textbf{MentalRoBERTa} \cite{ji2021mentalbertpubliclyavailablepretrained} by further training these models on large collections of mental health related texts from online platforms. The additional training helps the models to learn psychological terms, emotional expressions, and patterns in conversations that are common in mental health talks.

\par Similarly, Large Language Models (LLMs) have also recently shown strong performance in mental-health related tasks, where understanding context, emotion, and subtle language patterns is important. Recent work has also explored reasoning based prompting methods for mental health prediction. In particular, \textit{Avinash Patil and Amardeep Kour Gedhu} \cite{patil2025cognitive} studied the use of OpenAI's \textbf{o3-mini} model for mental-health text classification and evaluated prompting strategies such as zero shot, Chain of Thought (CoT), Few shot CoT, Self Consistency CoT, and Tree of Thought. Their work showed that prompt based reasoning can improve performance on several mental health datasets and highlighted the usefulness of step by step reasoning in mental health classification. Motivated by such prompting based approaches, we evaluate \textbf{MentalLLaMA} \cite{Yang_2024} in our work using zero shot and few shot CoT settings and compare its performance with our proposed multi task framework.

\section{Methodology}
\subsection{Problem Formulation}
The Core Fusion Network (CFN) is extended by jointly modeling Mental Health Condition classification alongside Emotion, Sentiment, and Topic classification as secondary tasks. Each secondary task provides additional contextual information that enriches the shared representation $h_i$, helping the model better classify mental health conditions.

Given a Reddit post $x_i$ from the SWMH dataset \cite{swmh}, the main objective is to predict four task labels: Mental Health Condition $y_i^{c}$, Emotion $y_i^{e}$, Sentiment $y_i^{s}$, and Topic $y_i^{t}$. Mental Health Condition is the primary task, while Emotion, Sentiment, and Topic act as secondary tasks that provide additional information about emotional state, sentiment polarity, and overall context of the post. The goal is to learn a shared representation $h_i$ that captures these relationships while predicting each task label, formulated as:

\begin{equation}
    f_\theta(x_i) \rightarrow (\hat{y}_i^{c}, \hat{y}_i^{e}, \hat{y}_i^{s}, \hat{y}_i^{t})
    \label{eq:multitaskcfn}
\end{equation}

where $f_\theta$ denotes the extended Core Fusion Network (CFN). In this study, $|Y^{c}| = 5$ (\textit{Depression}, \textit{Suicide Watch}, \textit{Anxiety}, \textit{Off My Chest}, and \textit{Bipolar}), $|Y^{e}| = 7$ (\textit{Joy}, \textit{Sadness}, \textit{Anger}, \textit{Fear}, \textit{Disgust}, \textit{Surprise}, \textit{Neutral}), $|Y^{s}| = 3$ (\textit{Positive}, \textit{Negative}, \textit{Neutral}), and $|Y^{t}| = 31$ (topic categories derived from CounselChat dataset \cite{bertagnolli2020counsel}).

The key challenge is that the secondary task labels ($y_i^{e}$, $y_i^{s}$, and $y_i^{t}$) are generated by pre-trained models and may contain noise, which can propagate through the shared representation $h_i$ and reduce model performance. To address this challenge, this work proposes an Active Learning-based Stepwise Relabeling approach that identifies uncertain emotion and topic label predictions and progressively refines them through stepwise relabeling, improving label quality while preserving learned task representations across all four tasks.

\subsection{Label Generation Pipeline}\label{sec:label_gen}
The SuicideWatch and Mental Health (SWMH) dataset \cite{swmh} contains ground-truth labels only for primary mental health conditions. As shown in Figure~\ref{fig:pipeline_overview}(1), three secondary task label sets are generated to capture additional contextual signals across user posts. Emotion and sentiment labels are generated using pre-trained classifiers following Pathak et al. \cite{pathak2025}, whereas topic labels are extracted by mapping posts to CounselChat clinical Q\&A representations.

\begin{figure}[H]
\centering
\includegraphics[width = \linewidth]{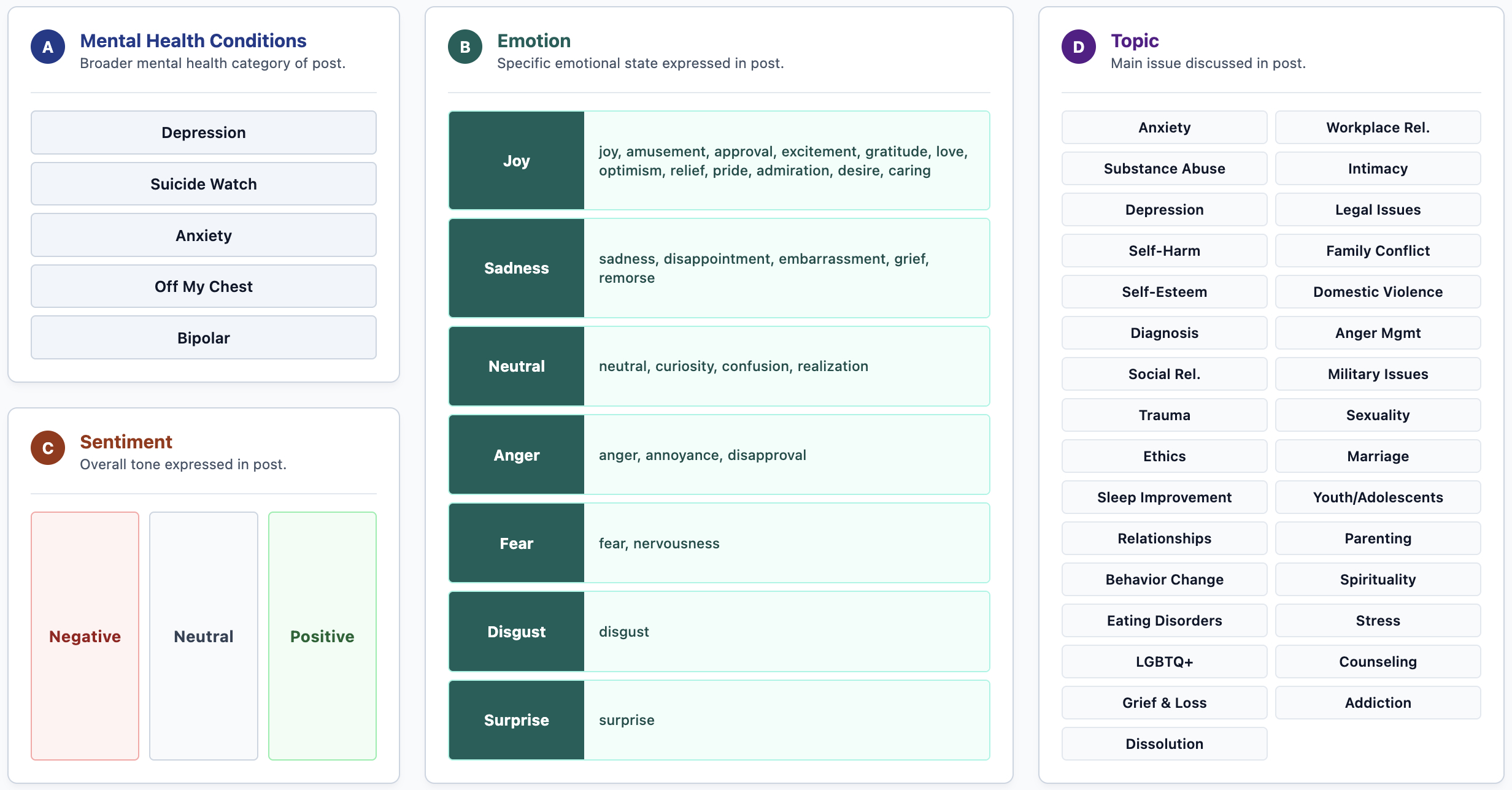}
\caption{\textbf{Classification task categories in the SWMH dataset.}}
\label{fig:Categories}
\end{figure}

\paragraph{Emotion Labels}
Emotion refers to the emotional state expressed in a specific post (e.g., \textit{sadness, fear}), capturing how an individual with a mental health condition can express different emotions across different posts. Emotion labels are generated using the BERT-GoEmotions model (\texttt{monologg/bert-base-cased-goemotions-original}) \cite{monologg_goemotions}, which classifies each post into 28 fine-grained categories from the GoEmotions dataset \cite{DBLP:journals/corr/abs-2005-00547}. The 28 emotion categories are mapped into 7 broader classes following Pathak et al. \cite{pathak2025}: Ekman's 6 basic emotions (\textit{Joy}, \textit{Sadness}, \textit{Anger}, \textit{Fear}, \textit{Disgust}, \textit{Surprise}) and a \textit{Neutral} class. This grouping adapts the official Ekman-based mapping from GoEmotions \cite{DBLP:journals/corr/abs-2005-00547}, where ambiguous categories (\textit{realization}, \textit{confusion}, and \textit{curiosity}) are assigned to \textit{Neutral} instead of \textit{Surprise}, as shown in Figure~\ref{fig:Categories}(B).

\paragraph{Sentiment Labels}
Sentiment measures the overall tone of the post. Sentiment labels are generated using the Twitter-RoBERTa model (\texttt{cardiffnlp/twitter-roberta-base-sentiment-latest}) \cite{loureiro-etal-2022-timelms}, which classifies each post into three categories: \textit{Positive}, \textit{Negative}, and \textit{Neutral}, as shown in Figure~\ref{fig:Categories}(C).

\paragraph{Topic Labels} Topic identifies the main subject discussed (e.g., \textit{family conflict, relationships, or identity}). Because individuals with the same mental health condition often experience distinct life stressors, topic classification provides a clinically meaningful signal. Labels are assigned using cosine similarity between SWMH post embeddings and CounselChat topic representations. SWMH posts and CounselChat question-and-answer texts are encoded into 768-dimensional embeddings using the \texttt{all-mpnet-base-v2} sentence transformer \cite{song2020mpnetmaskedpermutedpretraining}. A representative vector $\mathbf{v}_t$ for each of the 31 topics (shown in Figure~\ref{fig:Categories}(D)) is created by averaging the embeddings of all CounselChat text entries belonging to topic $t \in \{1, \dots, 31\}$. Each SWMH post $p$ is then assigned the topic with the highest cosine similarity score:

\begin{equation}
\mathrm{cosSim}(\mathbf{e}_p,\mathbf{v}_t)
=
\frac{\mathbf{e}_p \cdot \mathbf{v}_t}
{\|\mathbf{e}_p\| \, \|\mathbf{v}_t\|}
\label{eq:cosine_similarity}
\end{equation}

\begin{equation}
\mathrm{Topic}(p)
=
\arg\max_{t \in \{1,\ldots,31\}}
\mathrm{cosSim}(\mathbf{e}_p,\mathbf{v}_t)
\label{eq:topic_assignment}
\end{equation}

where $\mathbf{e}_p$ represents the embedding vector of SWMH post $p$, and $\mathbf{v}_t$ denotes the mean embedding of all CounselChat Q\&A pairs for topic $t$.

\subsection{Core Fusion Network}\label{sec:cfn}
\subsubsection{Overview}
The proposed Core Fusion Network (CFN), shown in Figure~\ref{fig:pipeline_overview}(2), consists of three primary components: a domain-specific shared feature extractor (\texttt{MentalBERT} / \texttt{MentalRoBERTa}) \cite{ji2021mentalbertpubliclyavailablepretrained}, four task-specific models, and a central fusion model that captures cross-task dependencies for final classification. Following initial training, noisy emotion and topic labels are iteratively refined using an Active Learning-based Stepwise Relabeling pipeline before retraining.

\subsubsection{Feature Extractor}
Two domain-specific transformer encoders are evaluated: MentalBERT (\texttt{mental-bert-base-uncased}) and MentalRoBERTa (\texttt{mental-roberta-base}) \cite{ji2021mentalbertpubliclyavailablepretrained}, both pre-trained on mental health-related Reddit text to capture domain-specific linguistic patterns.

Given a Reddit post $x_i$, the encoder produces contextual token embeddings

\begin{equation}
    \mathbf{H}_i = \operatorname{Encoder}(x_i) \in \mathbb{R}^{n \times d}
    \label{eq:featureextractor}
\end{equation}

where $n = 512$ is the sequence length after padding or truncation, and $d = 768$ is the hidden dimension of the encoder. $\mathbf{H}_i$ is passed as shared input to four task-specific models.

\subsubsection{Task-Specific Models} Four independent task-specific models are created for Mental Health Condition, Emotion, Sentiment, and Topic. For task $k \in \{c,e,s,t\}$, the shared embeddings from the encoder $\mathbf{H}_i$ pass through an LSTM layer. From the LSTM output, two parallel intermediate representations are derived: $I_1^{(k)}$, computed via a ReLU-activated linear transformation, and $I_2^{(k)}$, computed by passing the LSTM features through a multi-head attention layer followed by a linear transformation. Finally, the preliminary task-specific prediction $\tilde{y}_i^{(k)}$ is obtained by applying dropout and a linear classifier to the last timestep of the multi-head attention output.

\subsubsection{Central Fusion Model}
The four task-specific predictions are concatenated into $\mathbf{X}_c$ = $[\tilde{y}_i^{(c)}, \tilde{y}_i^{(e)}, \tilde{y}_i^{(s)}, \tilde{y}_i^{(t)}]$, and processed by a central fusion model consisting of an LSTM layer followed by an attention layer, which captures cross-task dependencies. The last timestep of this central attention output forms the shared representation $h_i$.

\subsubsection{Fully Connected Layer and Output}
For each task $k$, the shared representation $h_i$ is concatenated with the intermediate layers $I_1^{(k)}$ and $I_2^{(k)}$ generated by the corresponding task-specific model and the result is passed through a fully connected layer, consisting of a linear layer, a ReLU activation function, a dropout layer ($p=0.5$), and a final linear layer to produce the final prediction $\hat{y}_i^{(k)}$ for each task.

\subsubsection{Loss Function}
To address the class imbalance in the dataset, each task is trained using a class-weighted categorical cross-entropy loss:
\begin{equation}
L_k = -\sum_{j} w_j^{(k)}\,y_{ij}^{(k)}
\log\!\left(\hat{y}_{ij}^{(k)}\right)
\label{eq:class_weight_crossentropy_loss}
\end{equation}
where $w_j^{(k)}$ is the inverse frequency class weight for class $j$ in each task $k \in \{c,e,s,t\}$, $y_{ij}^{(k)}$ is the true label for class $j$, and $\hat{y}_{ij}^{(k)}$ is the predicted probability for class $j$. The total loss is a weighted combination of all four task losses:
\begin{equation}
    L_T = \alpha L_c + \beta L_e + \gamma L_s + \delta L_t
    \label{eq:total_loss}
\end{equation}
with $\alpha = 0.45$, $\beta = 0.25$, $\gamma = 0.15$, $\delta = 0.15$, prioritizing Mental Health Condition as primary task, while Emotion, Sentiment, and Topic are weighted as secondary tasks.

\subsection{Active Learning-based Stepwise Relabeling}
The trained CFN serves as the baseline model for the proposed Active Learning-based Stepwise Relabeling approach (Figure~\ref{fig:pipeline_overview}(3)). To address the noise present in the emotion and topic labels generated by pre-trained models, this approach specifically targets these two tasks for label refinement. Sentiment labels, which achieve consistently higher predictive accuracy, are excluded from relabeling.

\subsubsection{Uncertainty Estimation}
For each training sample $x_i$, the baseline CFN produces predicted probability distributions for emotion and topic. The uncertainty of each task $k \in \{e, t\}$ is calculated using entropy:

\begin{equation}
    H^{k}(x_i) = -\frac{1}{\ln(|Y^{k}|)}\sum_{j=1}^{|Y^{k}|} p(y_i^{k}=j \mid x_i) \ln p(y_i^{k}=j \mid x_i)
    \label{eq:entropy}
\end{equation}

where $j$ is the class index, $p(y_i^{k}=j \mid x_i)$ is the predicted probability assigned to class $j$ by the CFN, and $|Y^{k}|$ is the total number of classes for task $k$ ($|Y^{e}| = 7$ for emotion and $|Y^{t}| = 31$ for topic). The resulting normalized entropy $H^{k}(x_i) \in [0,1]$ measures prediction uncertainty, where higher values indicate greater uncertainty.

\subsubsection{Relabeling Pipeline}
Samples selected for relabeling are processed through task-specific pre-trained models. Emotion labels are updated using \textit{SamLowe/roberta-base-go\_emotions} \cite{sam_lowe_2024}; the resulting 28 GoEmotions categories are mapped to the 7 emotion categories using the same grouping described in Section \ref{sec:label_gen}, while topic labels are updated via \textit{facebook/bart-large-mnli} zero-shot classification \cite{lewis2019bartdenoisingsequencetosequencepretraining}. A predicted label is accepted only if its confidence score satisfies $\text{score} \geq \tau$, where $\tau = 0.70$; otherwise, the original label is retained.

\subsubsection{Relabeling Strategies}

Three uncertainty-based relabeling strategies are explored:
\begin{enumerate}
    \item \textbf{Medium Uncertainty:} Training samples with prediction uncertainty between the $40^{th}$ and $70^{th}$ percentiles are identified as medium uncertainty samples. From this pool, the top $3{,}000$ samples ranked by prediction uncertainty are selected for relabeling. The CFN is then retrained from scratch on the updated training set in a single stage.
    \item \textbf{High Uncertainty:} The top $3{,}000$ training samples ranked by prediction uncertainty are identified as highly uncertain and selected for relabeling. The CFN is then retrained from scratch on the updated training set in a single stage.
    \item \textbf{Active Learning-based Stepwise Relabeling:} This strategy combines both uncertainty levels in a sequential two-stage process. In \textit{Stage 1}, the medium uncertainty samples are relabeled, and the CFN is retrained from scratch at a learning rate of $1\times 10^{-5}$. In \textit{Stage 2}, the prediction uncertainty is recomputed on the updated training set of Stage 1, and the top $3{,}000$ most uncertain samples are relabeled. The Stage 1 model is then fine-tuned on this further updated training set at a reduced learning rate of $5 \times 10^{-6}$, allowing the model to refine predictions on the most uncertain samples while preserving the task representations learned in Stage 1.
\end{enumerate}

\section{Experiments}
\subsection{Setup}
\subsubsection{Dataset}
This study uses two datasets: the SuicideWatch and Mental Health (SWMH) dataset \cite{swmh} for model training and evaluation, and the CounselChat dataset \cite{bertagnolli2020counsel} for deriving topic labels.

\paragraph{Suicide Watch Mental Health (SWMH) Dataset}
 The SWMH dataset \cite{swmh} contains 54,412 Reddit posts collected through the official Reddit API from five mental health subreddits: \textit{r/Depression}, \textit{r/Suicide Watch}, \textit{r/Anxiety}, \textit{r/Off My Chest}, and \textit{r/Bipolar}. The dataset includes 34,823 training posts, 8,706 validation posts, and 10,883 test posts. Following Ji et al. \cite{swmh}, the subreddit from which each post was collected is treated as its mental health condition label.

 \paragraph{CounselChat Dataset} The CounselChat dataset \cite{bertagnolli2020counsel} was used to derive topic labels for the SWMH \cite{swmh} dataset. The dataset consists of 2,775 question-and-answer pairs where licensed therapists answer mental health related questions posted by users, covering 31 unique topics including depression, anxiety, relationships, trauma, family conflict, and self-harm. 

 \paragraph{Text Preprocessing} All text across SWMH and CounselChat is cleaned using the same pipeline: missing values are filled with empty strings, HTML tags, URLs, and non-alphabetic characters are removed, whitespace is normalized, and the text is lowercased. For CounselChat, the question and answer texts are concatenated into a single string before cleaning, producing one combined text per Q\&A pair for topic vector generation.

\begin{figure}[H]
\centering
\includegraphics[width=\textwidth]{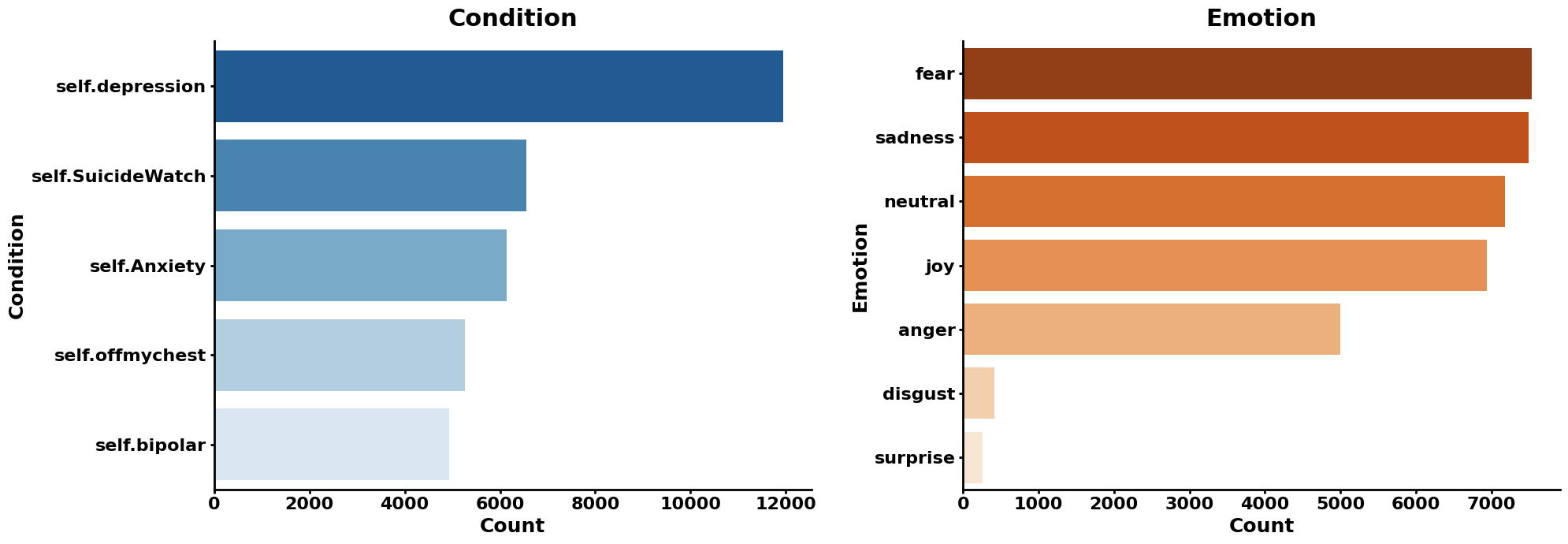}
\caption{\textbf{Mental Health Condition and Emotion label distributions in the SWMH training set.}}
\label{fig:cond_emotion_dist}
\end{figure}

\begin{figure}[H]
\centering
\includegraphics[width=\textwidth]{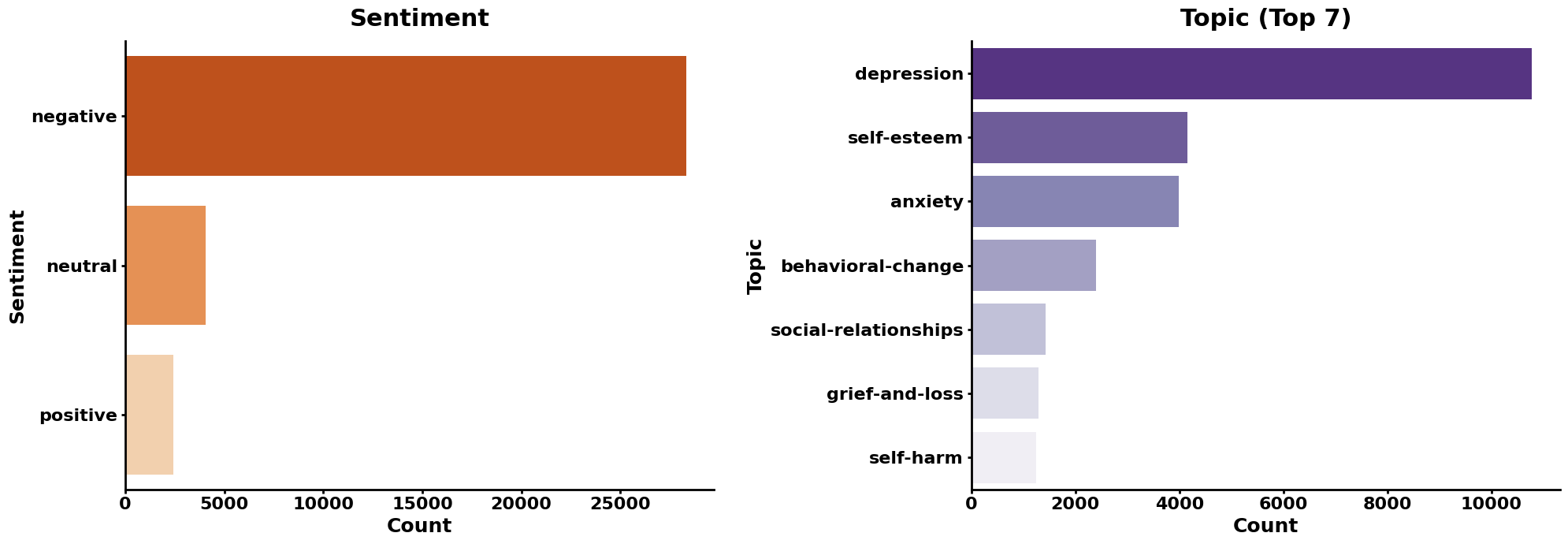}
\caption{\textbf{Sentiment and Topic label distributions in the SWMH training set.}}
\label{fig:sent_topic_dist}
\end{figure}

\paragraph{Label Distribution} Figure~\ref{fig:cond_emotion_dist} and~\ref{fig:sent_topic_dist}  show the label distributions for all four tasks in the SWMH \cite{swmh} training set, the ground truth mental health condition label and the three auxiliary tasks for emotion, sentiment, and topic. The class imbalance varies by task, which can push the model to favor the majority classes.

\begin{enumerate}
    \item The condition labels are moderately imbalanced, with \textit{Depression} having the most samples (11,940) and \textit{Bipolar} the fewest (4,932). The remaining three classes are more evenly distributed.

    \item The emotion labels are imbalanced, with fear (7,534), sadness (7,493), and neutral (7,181) having the most samples, while disgust (413) and surprise (261) appear rarely. Posts discussing ongoing mental health challenges often convey sustained emotions such as fear and sadness rather than transient reactions such as surprise or disgust. The grouping of 28 GoEmotions categories into 7 broader emotion classes adds to this imbalance, as multiple fine-grained emotions map onto fear and sadness alone.
    \item The sentiment labels are highly imbalanced, with 28,322 negative posts, more than neutral or positive posts. This negative skew is expected, since people usually post in the mental health subreddits when they are having a hard time.
    \item Depression is the most common topic with 10,787 posts, followed by Self-esteem with 4,158 posts and anxiety with 3,995. In contrast, topics such as marriage (56), children-adolescents (47), and professional-ethics (43) appear in fewer than 100 posts.
\end{enumerate}

\begin{figure}[H]
\centering
\includegraphics[width=\textwidth]{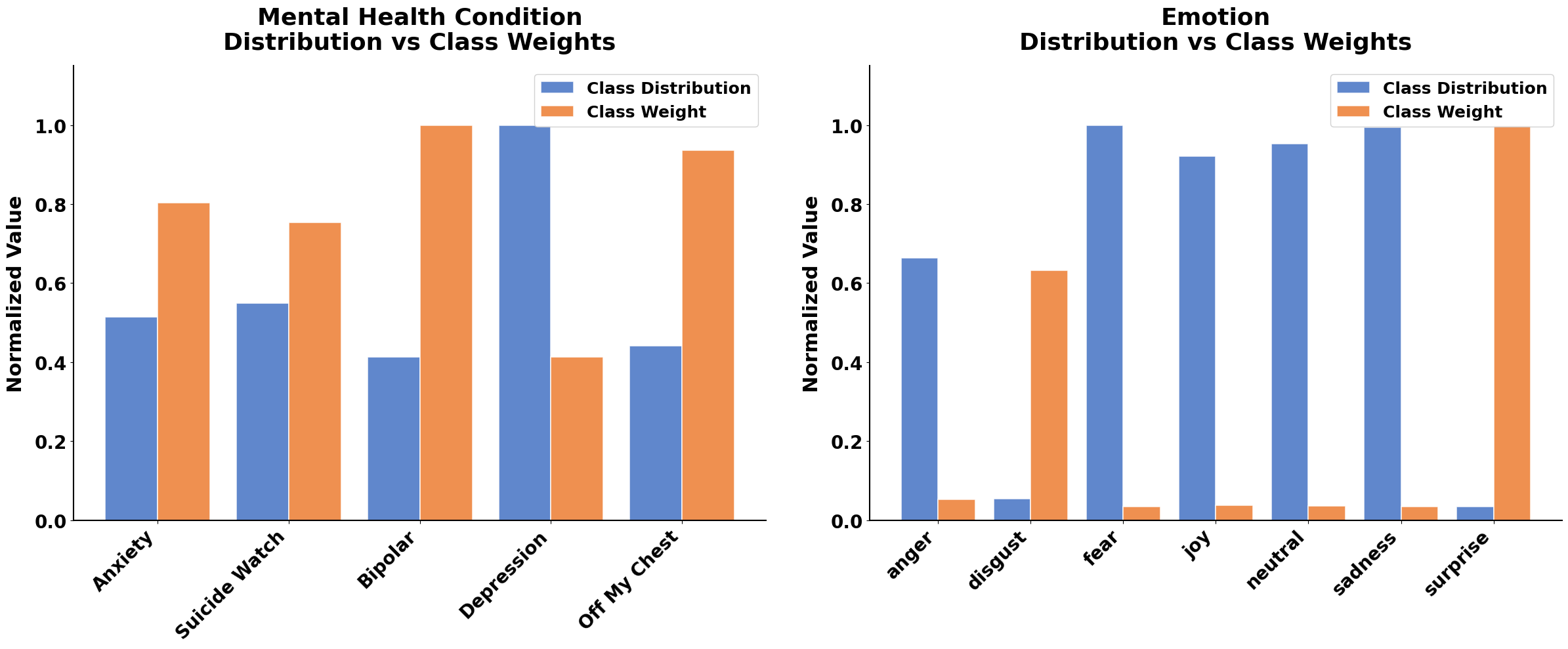}
\caption{\textbf{Class distribution vs. class weights for Mental Health Condition and Emotion}}
\label{fig:cond_emotion_weights}
\end{figure}

\begin{figure}[H]
\centering
\includegraphics[width=\textwidth]{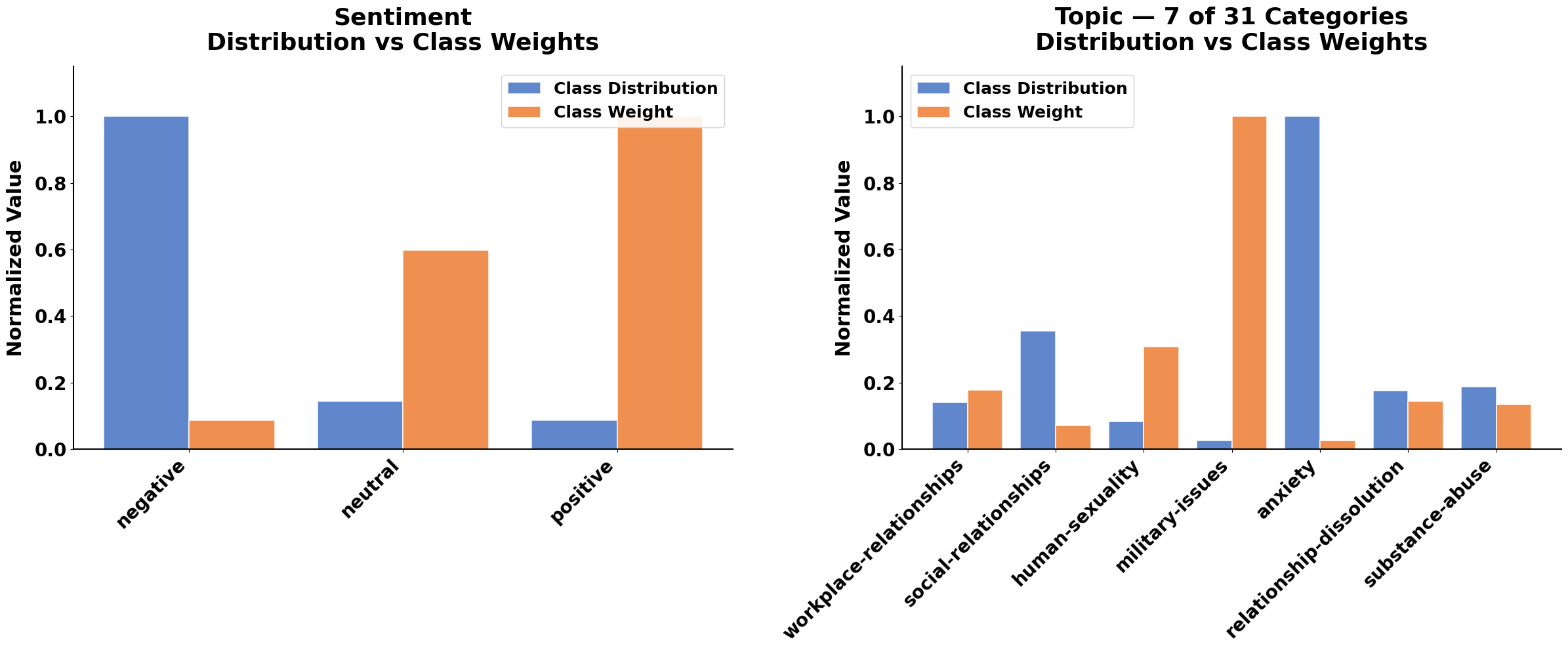}
\caption{\textbf{Class distribution vs. class weights for Sentiment and Topic}}
\label{fig:sent_topic_weights}
\end{figure}

\paragraph{Class Imbalance Handling}
To prevent the model from favoring majority classes, inverse-frequency class weights are computed for each task and applied to the categorical cross-entropy loss during training. As Figure~\ref{fig:cond_emotion_weights} and ~\ref{fig:sent_topic_weights}  show, rare classes such as \textit{Bipolar} in mental health conditions, \textit{Surprise} in emotion, \textit{Positive} in sentiment, and \textit{Professional-ethics} in topic receive higher weights, while frequent classes such as \textit{Depression}, \textit{neutral}, \textit{negative}, and \textit{depression} receive lower weights.

\subsubsection{Baselines}
The extended CFN is compared against the following baselines using MentalRoBERTa. 
\begin{enumerate}
    \item The Multi-task Fully-Shared Network (Mt-FnS) uses a single shared MentalRoBERTa encoder, LSTM and multi-head attention layer with four independent task classifiers applied to the shared representation.
    \item The Three-task CFN \cite{pathak2025} is the original CFN jointly modeling Mental Health Condition, Emotion and Sentiment tasks without Topic.
    \item MentalLLaMA-7B \cite{Yang_2024} is evaluated with 4-bit quantization (QLoRA) on 750 samples from the SWMH \cite{swmh} test set, with 150 samples from each mental health condition class, using zero-shot and few-shot prompting to predict all four tasks simultaneously.
\end{enumerate}

\subsubsection{Metrics}
All results are evaluated on the held-out test set of 10,883 samples using Accuracy, Precision, Recall, and F1-score.

\subsection{Implementation Details}
Table~\ref{tab:experimental_setup} summarizes the training configuration used across all experiments. The best model checkpoint is selected based on the weighted validation score computed as:

\begin{equation}
    \text{val\_score} = 0.45 \cdot \text{acc}_c + 0.25 \cdot \text{acc}_e + 0.15 \cdot \text{acc}_s + 0.15 \cdot \text{acc}_t
\end{equation}

where $\text{acc}_c$, $\text{acc}_e$, $\text{acc}_s$, $\text{acc}_t$ are the validation accuracies for mental health condition, emotion, sentiment, and topic classification. The weighted score reflects the importance of each task, with the highest weight assigned to the Mental Health Condition classification ($\alpha = 0.45$), the primary task.

\begin{table}[h] 
\captionsetup{justification=centering}
\caption{Experimental Configuration used for training the Core Fusion Network (CFN)}\label{tab:experimental_setup}%
\begin{tabular}{@{}ll@{}} 
\toprule 
\textbf{Parameter} & \textbf{Value} \\ 
\midrule
Optimizer & AdamW \\ 
Learning Rate & $1 \times 10^{-5}$ \\ 
Weight Decay & $0.01$ \\ 
Scheduler & Linear warmup-decay (10\% warmup) \\
Gradient Clipping & $1.0$ \\ 
Batch Size & $32$ \\ 
Epochs & $50$ \\ 
Early Stopping & $5$ epochs \\ 
Random Seed & $42$ \\ 
Max Sequence Length & $512$ tokens \\ 
Hidden Size & $256$ \\ 
Dropout & $0.5$ \\ 
Attention Heads & 4 \\ 
\end{tabular}
\end{table}
\newpage
\subsection{Results and Analysis}
The extended Core Fusion Network (CFN) with the proposed Active Learning-based Stepwise Relabeling approach is evaluated on the SWMH dataset \cite{swmh} across two domain-specific encoders, MentalBERT and MentalRoBERTa \cite{ji2021mentalbertpubliclyavailablepretrained}. All results are reported on the held-out test set of 10,883 samples using accuracy, precision, recall, and F1-score. The analysis is structured around three research questions: whether the proposed Stepwise Relabeling approach improves over the extended CFN trained on the original labels, referred to as the baseline CFN (RQ1), how the extended CFN compares against simpler multi-task baselines (RQ2), and which Active Learning strategy contributes most to overall performance (RQ3).

\subsubsection{Emotion Classification: GoEmotions(28 categories) vs GoEmotions mapped to 7 categories}

The CFN with MentalRoBERTa is evaluated on all 28 GoEmotions categories directly to assess whether fine-grained emotion labels improve classification. For comparison, results with GoEmotions mapped to 7 categories as described in Section~\ref{sec:label_gen} are also reported. Table~\ref{tab:goemotions28} shows the emotion classification results for both configurations for the Baseline CFN and Stepwise Relabeling Stage 2.

\begin{table*}[!ht]
\centering
\caption{Emotion accuracy comparison: 28 GoEmotions categories vs. 7 grouped categories for the Baseline CFN and Stepwise Relabeling Stage 2 (MentalRoBERTa).}
\label{tab:goemotions28}
\resizebox{\linewidth}{!}{%
\begin{tabular}{lcccccccc}
\toprule
& \multicolumn{4}{c}{\textbf{Baseline CFN}}
& \multicolumn{4}{c}{\textbf{Stepwise Relabeling Stage 2}} \\
\cmidrule(lr){2-5} \cmidrule(lr){6-9}
\textbf{Categories} & Accuracy & Precision & Recall & F1 & Accuracy & Precision & Recall & F1 \\
\midrule
28 GoEmotions & 0.5740 & 0.5757 & 0.5740 & 0.5723 & 0.3292 & 0.3099 & 0.3292 & 0.2309 \\
7 Grouped Categories  & \textbf{0.7032} & \textbf{0.7039} & \textbf{0.7032} & \textbf{0.7030} & \textbf{0.7083} & \textbf{0.7075} & \textbf{0.7083} & \textbf{0.7066} \\
\bottomrule
\end{tabular}%
}
\end{table*}

The 7-class mapping consistently outperforms the 28 GoEmotions categories in both Baseline and Stepwise Relabeling Stage 2, achieving an emotion accuracy of 0.7032 at baseline and 0.7083 under Stepwise Relabeling Stage 2. This suggests that relabeling across 28 fine-grained emotions introduces more noise than it removes. Mapping GoEmotions into 7 broader classes is necessary for reliable emotion classification in mental health text.

\subsubsection{Main Results}
Table~\ref{tab:main_results} addresses RQ1 by reporting the performance of the CFN with and without Active Learning-based Stepwise Relabeling across both encoders. Stepwise Relabeling improves all four tasks over the baseline CFN, with the largest improvements on Topic (+1.61\% MentalBERT, +1.57\% MentalRoBERTa), followed by Sentiment (+0.40\% MentalBERT, +0.35\% MentalRoBERTa) and Emotion (+0.09\% MentalBERT, +0.51\% MentalRoBERTa). Mental Health Condition accuracy also improves slightly in both encoders under Stepwise Relabeling Stage 2 (+0.24\% MentalBERT, +0.09\% MentalRoBERTa). MentalRoBERTa outperforms MentalBERT on Mental Health Condition, Sentiment, and Topic, and is used as the primary encoder in all analyses.

\begin{table*}[!ht]
\centering
\caption{Main Results: Baseline CFN vs. 
Active Learning-based Stepwise Relabeling (Stage 2) for MentalBERT and MentalRoBERTa encoders. Best Accuracy per task is 
in \textbf{bold}.}
\label{tab:main_results}
\resizebox{\linewidth}{!}{%
\begin{tabular}{lcccccccccccccccc}
\toprule
& \multicolumn{8}{c}{\textbf{Core Fusion Network - MentalBERT}} 
& \multicolumn{8}{c}{\textbf{Core Fusion Network - MentalRoBERTa}} \\

& \multicolumn{4}{c}{\textit{\textbf{Baseline}}} 
& \multicolumn{4}{c}{\textit{\textbf{Stepwise Relabeling Stage 2}}} 
& \multicolumn{4}{c}{\textit{\textbf{Baseline}}}
& \multicolumn{4}{c}{\textit{\textbf{Stepwise Relabeling Stage 2}}} \\
\cmidrule(lr){2-5} \cmidrule(lr){6-9}
\cmidrule(lr){10-13} \cmidrule(lr){14-17}
\textbf{Task} 
& Accuracy & Precision & Recall & F1  
& Accuracy & Precision & Recall & F1  
& Accuracy & Precision & Recall & F1  
& Accuracy & Precision & Recall & F1  \\
\midrule
Condition
& 0.6846 & 0.6914 & 0.6846 & 0.6847
& 0.6870 & 0.6907 & 0.6870 & 0.6880
& 0.6965 & 0.7003 & 0.6965 & 0.6974
& \textbf{0.6974} & 0.6995 & 0.6974 & 0.6980 \\
Emotion
& 0.7188 & 0.7182 & 0.7188 & 0.7180
& \textbf{0.7197} & 0.7191 & 0.7197 & 0.7190
& 0.7032 & 0.7039 & 0.7032 & 0.7030
& 0.7083 & 0.7075 & 0.7083 & 0.7066 \\
Sentiment
& 0.9003 & 0.9009 & 0.9003 & 0.9005
& 0.9043 & 0.9044 & 0.9043 & 0.9043
& 0.9168 & 0.9160 & 0.9168 & 0.9164
& \textbf{0.9203} & 0.9174 & 0.9203 & 0.9184 \\
Topic
& 0.5637 & 0.6136 & 0.5637 & 0.5741
& 0.5798 & 0.6194 & 0.5798 & 0.5877
& 0.5940 & 0.6407 & 0.5940 & 0.6045
& \textbf{0.6097} & 0.6483 & 0.6097 & 0.6190 \\
\bottomrule
\end{tabular}%
}
\end{table*}

\subsubsection{Comparison with Multi-Task Baselines}
To address RQ2, the extended CFN with Stepwise Relabeling stage 2 is compared against two baselines, Multi-task Fully-Shared Network (Mt-FnS) and three-task CFN using MentalRoBERTa.

Table~\ref{tab:comparison} shows that the proposed model outperforms Mt-FnS on all four tasks, including Emotion (+0.22\%) and Topic (+2.01\%), as well as Mental Health Condition (+0.79\%) and Sentiment (+1.14\%). Against the three-task CFN, the proposed model achieves comparable Mental Health Condition accuracy (0.6974 vs.\ 0.6970) and higher Emotion (+3.13\%) and Sentiment (+0.33\%) accuracy. Adding Topic as a fourth secondary task does not degrade the original three tasks, as the extended CFN baseline matches the three-task CFN on Mental Health Condition and Sentiment while improving Emotion from 0.6770 to 0.7032. Overall, the proposed model achieves the highest accuracy of all four tasks among the models in Table~\ref{tab:comparison}, with the largest gain over Mt-FnS on Topic.

\begin{table*}[!ht]
\centering
\caption{Comparison of the Extended CFN with Active Learning-based Stepwise Relabeling against multi-task baselines using MentalRoBERTa. Best Accuracy per task is in \textbf{bold}. The Three-Task CFN does not include Topic classification.}
\label{tab:comparison}
\resizebox{\linewidth}{!}{%
\begin{tabular}{lcccccccccccccccc}
\toprule
& \multicolumn{4}{c}{\textbf{Mt-FnS}} 
& \multicolumn{4}{c}{\textbf{Three-Task CFN}} 
& \multicolumn{4}{c}{\textbf{Extended CFN}} 
& \multicolumn{4}{c}{\textbf{Extended CFN}} \\
& \multicolumn{4}{c}{\textit{\textbf{Baseline}}} 
& \multicolumn{4}{c}{\textit{\textbf{Baseline}}} 
& \multicolumn{4}{c}{\textit{\textbf{Baseline}}} 
& \multicolumn{4}{c}{\textit{\textbf{+ Stepwise Relabeling Stage 2}}} \\
\cmidrule(lr){2-5} \cmidrule(lr){6-9}
\cmidrule(lr){10-13} \cmidrule(lr){14-17}
\textbf{Task} 
& Accuracy & Precision & Recall & F1 
& Accuracy & Precision & Recall & F1 
& Accuracy & Precision & Recall & F1  
& Accuracy & Precision & Recall & F1  \\
\midrule
Condition
& 0.6895 & 0.6953 & 0.6895 & 0.6902
& 0.6970 & 0.6980 & 0.6970 & 0.6960
& 0.6965 & 0.7003 & 0.6965 & 0.6974
& \textbf{0.6974} & 0.6995 & 0.6974 & 0.6980 \\
Emotion
& 0.7061 & 0.7064 & 0.7061 & 0.7043
& 0.6770 & 0.6780 & 0.6770 & 0.6760
& 0.7032 & 0.7039 & 0.7032 & 0.7030
& \textbf{0.7083} & 0.7075 & 0.7083 & 0.7066 \\
Sentiment
& 0.9089 & 0.9112 & 0.9089 & 0.9100
& 0.9170 & 0.9190 & 0.9170 & 0.9180
& 0.9168 & 0.9160 & 0.9168 & 0.9164
& \textbf{0.9203} & 0.9174 & 0.9203 & 0.9184 \\
Topic
& 0.5896 & 0.6316 & 0.5896 & 0.5984
& --- & --- & --- & ---
& 0.5940 & 0.6407 & 0.5940 & 0.6045
& \textbf{0.6097} & 0.6483 & 0.6097 & 0.6190 \\
\bottomrule
\end{tabular}}
\end{table*}
\newpage
Table~\ref{tab:stepwise_comparison} compares the effect of Active Learning-based Stepwise Relabeling Stage 2 on the extended CFN and the three-task CFN. Each architecture improves over its own baseline: the extended CFN on all four tasks (Table~\ref{tab:main_results}), and the three-task CFN on Mental Health Condition (0.697 to 0.7058), Emotion (0.677 to 0.7167), and Sentiment (0.917 to 0.9177) suggesting that the approach is not specific to the extended CFN. Comparing the two architectures after Stage 2, the three-task CFN reaches higher Mental Health Condition and Emotion accuracy, while the extended CFN achieves higher Sentiment accuracy (0.9203 vs.\ 0.9177) and additionally predicts Topic.

\begin{table*}[!ht]
\centering
\caption{Effect of Active Learning-based Stepwise Relabeling Stage 2 on the 
Extended CFN and Three-Task CFN using MentalRoBERTa. Best Accuracy per task is in 
\textbf{bold}. The Three-Task CFN does not include Topic classification.}
\label{tab:stepwise_comparison}
\resizebox{\linewidth}{!}{%
\begin{tabular}{lcccccccc}
\toprule
& \multicolumn{4}{c}{\textbf{Extended CFN}}
& \multicolumn{4}{c}{\textbf{Three-Task CFN}} \\
& \multicolumn{4}{c}{\textit{\textbf{+ Stepwise Relabeling Stage 2}}}
& \multicolumn{4}{c}{\textit{\textbf{+ Stepwise Relabeling Stage 2}}} \\
\cmidrule(lr){2-5} \cmidrule(lr){6-9}
\textbf{Task}
& Accuracy & Precision & Recall & F1
& Accuracy & Precision & Recall & F1 \\
\midrule
Condition
& 0.6974 & 0.6995 & 0.6974 & 0.6980
& \textbf{0.7058} & 0.7075 & 0.7058 & 0.7055 \\
Emotion
& 0.7083 & 0.7075 & 0.7083 & 0.7066
& \textbf{0.7167} & 0.7186 & 0.7167 & 0.7168 \\
Sentiment
& \textbf{0.9203} & 0.9174 & 0.9203 & 0.9184
& 0.9177 & 0.9210 & 0.9177 & 0.9191 \\
Topic
& \textbf{0.6097} & 0.6483 & 0.6097 & 0.6190
& --- & --- & --- & --- \\
\bottomrule
\end{tabular}%
}
\end{table*}

\subsubsection{Comparison with LLM Prompting}
MentalLLaMA-7B\cite{Yang_2024} with 4-bit quantization (QLoRA) is evaluated on 750 stratified samples from the SWMH\cite{swmh} test set, with 150 samples per mental health condition class, using zero-shot and few-shot prompting to predict Mental Health Condition, Emotion, Sentiment, and Topic for each post. As shown in Table~\ref{tab:llm_comparison}, the CFN outperforms MentalLLaMA-7B on Mental Health Condition, Emotion, and Sentiment Classification tasks, demonstrating that a fine-tuned multi-task CFN model outperforms zero-shot and few-shot LLM prompting. Topic accuracy is lower in this subset because the class distribution is imbalanced and 3 of 31 topic classes are missing entirely.

\begin{table}[!ht]
\centering
\small
\caption{Comparison of CFN (MentalRoBERTa) against MentalLLaMA-7B on a stratified subset of 750 samples. Best results per task are in \textbf{bold}.}
\label{tab:llm_comparison}
\begin{tabular}{lcccc}
\toprule
& \multicolumn{2}{c}{\textbf{CFN (MentalRoBERTa)}} & \multicolumn{2}{c}{\textbf{MentalLLaMA-7B (QLoRA)}} \\
\cmidrule(lr){2-3} \cmidrule(lr){4-5}
\textbf{Task} & Accuracy & F1 & Accuracy (Zero-Shot) & Accuracy (Few-Shot) \\
\midrule
Condition & \textbf{0.7400} & \textbf{0.7390} & 0.5333 & 0.5507 \\
Emotion   & \textbf{0.6427} & \textbf{0.6384} & 0.2800 & 0.3160 \\
Sentiment & \textbf{0.9027} & \textbf{0.9059} & 0.8333 & 0.7813 \\
Topic     & 0.1747          & 0.1578          & \textbf{0.3053} & \textbf{0.4253} \\
\bottomrule
\end{tabular}
\end{table}
\newpage
\subsection{Ablation Study}
Table~\ref{tab:ablation_study} addresses RQ3 by evaluating how each Active Learning strategy contributes to overall performance, using MentalRoBERTa as the encoder. For comparison, Random Sampling selects 3,000 training samples at random and relabels them using the same pipeline and confidence threshold. No single-stage strategy improves all four tasks. High Uncertainty sampling lowers Topic accuracy from 0.5940 to 0.5709, and Mental Health Condition accuracy also drops slightly to 0.6933. Medium Uncertainty sampling gives the best Emotion accuracy of any strategy (0.7106), though Mental Health Condition accuracy falls marginally to 0.6959. Random Sampling relabels the same number of samples as the Medium and High Uncertainty strategies, yet it has the lowest Mental Health Condition, Emotion, and Topic accuracy of all strategies, which shows that the gains come from entropy-based sample selection rather than the number of relabeled samples. Both stages of the Active Learning-based Stepwise Relabeling approach improve all four tasks over the baseline CFN. Stepwise Relabeling Stage 1 improves Emotion by 0.62\% and Topic by 0.81\%, and Stepwise Relabeling Stage 2 achieves the best Mental Health Condition (0.6974), Sentiment (0.9203), and Topic (0.6097) accuracy. These results suggest that Active Learning-based Stepwise Relabeling gives more consistent gains than single-stage strategies. 

\begin{table*}[!ht]
\centering
\caption{Ablation Study: Comparison of Active Learning strategies on the extended CFN using MentalRoBERTa. Best Accuracy per task is in \textbf{bold}.}
\label{tab:ablation_study}
\resizebox{\linewidth}{!}{%
\begin{tabular}{lcccccccccccc}
\toprule
& \multicolumn{12}{c}{\textbf{Core Fusion Network (MentalRoBERTa)}} \\
\cmidrule(lr){2-13}
& \multicolumn{2}{c}{\textbf{Baseline}} 
& \multicolumn{2}{c}{\textbf{High Uncertainty}} 
& \multicolumn{2}{c}{\textbf{Medium Uncertainty}} 
& \multicolumn{2}{c}{\textbf{Random Sampling}} 
& \multicolumn{2}{c}{\textbf{Stepwise Relabeling Stage 1}} 
& \multicolumn{2}{c}{\textbf{Stepwise Relabeling Stage 2}} \\
\cmidrule(lr){2-3} \cmidrule(lr){4-5} \cmidrule(lr){6-7} \cmidrule(lr){8-9} \cmidrule(lr){10-11} \cmidrule(lr){12-13}
\textbf{Task} & Accuracy & F1 & Accuracy & F1 & Accuracy & F1 & Accuracy & F1 & Accuracy & F1 & Accuracy & F1 \\
\midrule
Condition & 0.6965 & 0.6974 & 0.6933 & 0.6938 & 0.6959 & 0.6971 & 0.6915 & 0.6928 & 0.6971 & 0.6975 & \textbf{0.6974} & \textbf{0.6980} \\
Emotion   & 0.7032 & 0.7030 & 0.7074 & 0.7056 & \textbf{0.7106} & \textbf{0.7094} & 0.6978 & 0.6983 & 0.7094 & 0.7073 & 0.7083 & 0.7066 \\
Sentiment & 0.9168 & 0.9164 & 0.9186 & 0.9179 & 0.9200 & 0.9181 & 0.9188 & 0.9187 & 0.9175 & 0.9173 & \textbf{0.9203} & \textbf{0.9184} \\
Topic     & 0.5940 & 0.6045 & 0.5709 & 0.5811 & 0.6019 & 0.6097 & 0.5575 & 0.5702 & 0.6021 & 0.6115 & \textbf{0.6097} & \textbf{0.6190} \\
\bottomrule
\end{tabular}}
\end{table*}

\subsubsection{Threshold Analysis}
To select the confidence threshold $\tau$, the Active Learning-based Stepwise Relabeling approach is evaluated at $\tau = 0.60$, $0.70$, and $0.80$ (Table~\ref{tab:threshold}). A threshold of $\tau = 0.70$ is the only setting at which Stepwise Relabeling Stage 2 improves all four tasks over the baseline CFN. A lower threshold of $\tau = 0.60$ accepts more relabels and gives the best Topic accuracy (0.6276), but Mental Health Condition accuracy falls below the baseline from 0.6965 to 0.6909. Since Mental Health Condition is the primary task, a threshold that lowers it below the baseline is not preferred. A higher threshold of $\tau = 0.80$ lowers Mental Health Condition, Sentiment, and Topic accuracy below the baseline, with Topic accuracy dropping from 0.5940 to 0.5391. Based on these results, $\tau = 0.70$ is used in all other experiments.

\begin{table*}[!ht]
\centering
\caption{Effect of confidence threshold $\tau$ on 
Active Learning-based Stepwise Relabeling using the CFN with MentalRoBERTa. Best Stage 2 results per task are in \textbf{bold}.}
\label{tab:threshold}
\resizebox{\linewidth}{!}{%
\begin{tabular}{lcccccccccccc}
\toprule
& \multicolumn{4}{c}{\textbf{$\tau = 0.60$}} 
& \multicolumn{4}{c}{\textbf{$\tau = 0.70$}} 
& \multicolumn{4}{c}{\textbf{$\tau = 0.80$}} \\
\cmidrule(lr){2-5} \cmidrule(lr){6-9} \cmidrule(lr){10-13}
& \multicolumn{12}{c}{\textit{\textbf{Core Fusion Network with Active Learning-based Stepwise Relabeling}}} \\

& \multicolumn{2}{c}{\textit{\textbf{Stage S1}}} 
& \multicolumn{2}{c}{\textit{\textbf{Stage S2}}} 
& \multicolumn{2}{c}{\textit{\textbf{Stage S1}}} 
& \multicolumn{2}{c}{\textit{\textbf{Stage S2}}} 
& \multicolumn{2}{c}{\textit{\textbf{Stage S1}}} 
& \multicolumn{2}{c}{\textit{\textbf{Stage S2}}} \\
\cmidrule(lr){2-3} \cmidrule(lr){4-5}
\cmidrule(lr){6-7} \cmidrule(lr){8-9}
\cmidrule(lr){10-11} \cmidrule(lr){12-13}
\textbf{Task} 
& Accuracy & F1 & Accuracy & F1 
& Accuracy & F1 & Accuracy & F1 
& Accuracy & F1 & Accuracy & F1 \\
\midrule
Condition 
& 0.6862 & 0.6879 & 0.6909 & 0.6915 
& 0.6971 & 0.6975 & \textbf{0.6974} & \textbf{0.6980} 
& 0.6918 & 0.6924 & 0.6954 & 0.6956 \\
Emotion   
& 0.7075 & 0.7053 & 0.7053 & 0.7037 
& 0.7094 & 0.7073 & \textbf{0.7083} & \textbf{0.7066} 
& 0.7104 & 0.7079 & 0.7071 & 0.7060 \\
Sentiment   
& 0.9199 & 0.9184 & 0.9194 & 0.9181 
& 0.9175 & 0.9173 & \textbf{0.9203} & \textbf{0.9184} 
& 0.9132 & 0.9154 & 0.9166 & 0.9176 \\
Topic   
& 0.6238 & 0.6329 & \textbf{0.6276} & \textbf{0.6347} 
& 0.6021 & 0.6115 & 0.6097 & 0.6190 
& 0.5389 & 0.5498 & 0.5391 & 0.5501 \\
\bottomrule
\end{tabular}%
}
\end{table*}

\subsection{Qualitative Analysis}
\begin{figure}[H]
\centering
\begin{subfigure}[t]{\textwidth}
    \centering
    \includegraphics[width=\textwidth]{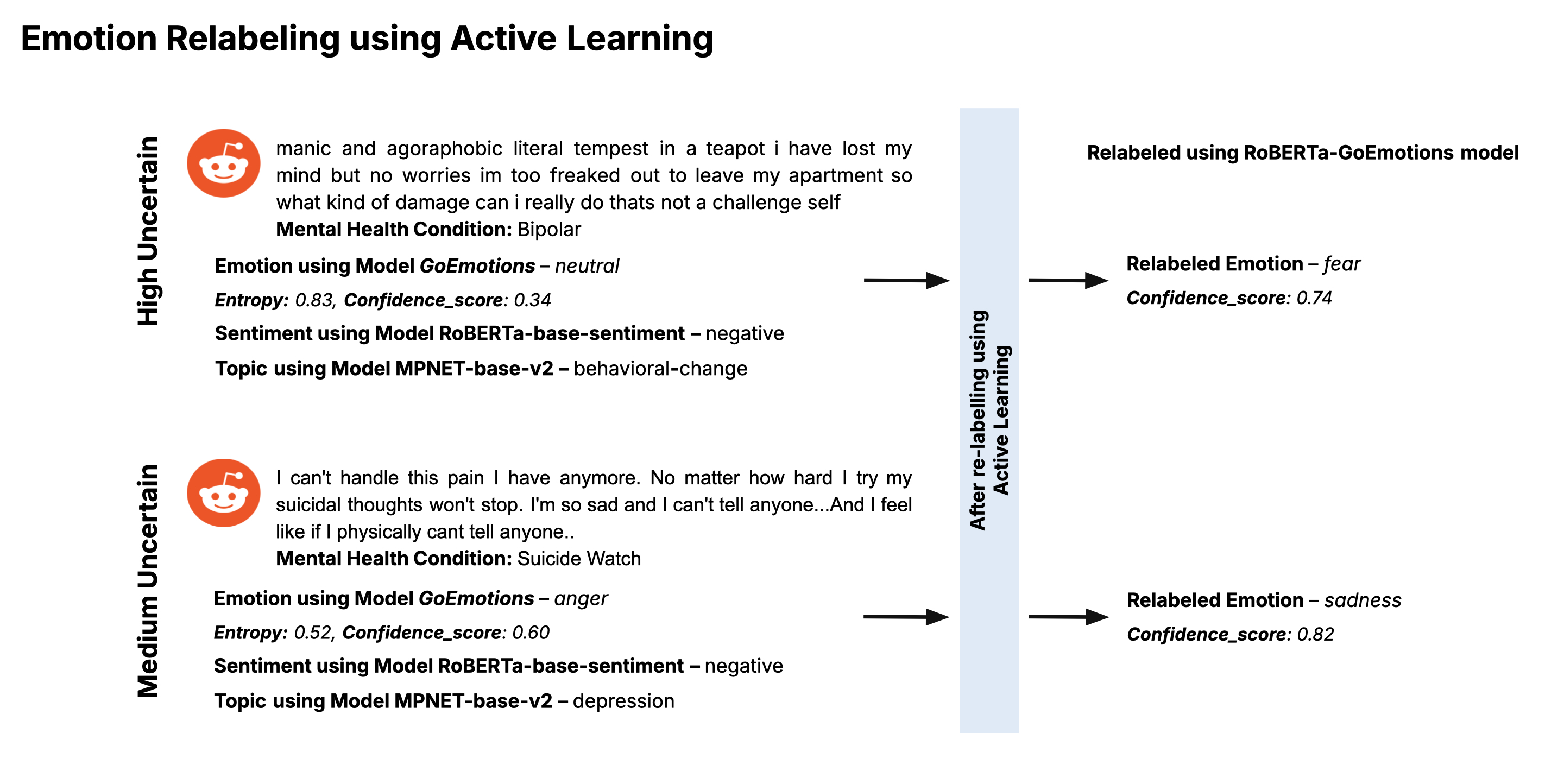}
    \caption{Emotion relabeling using Medium and High Uncertainty sampling}
    \label{fig:emo_relabel_hm}
\end{subfigure}

\vspace{0.5cm}

\begin{subfigure}[t]{\textwidth}
    \centering
    \includegraphics[width=\textwidth]{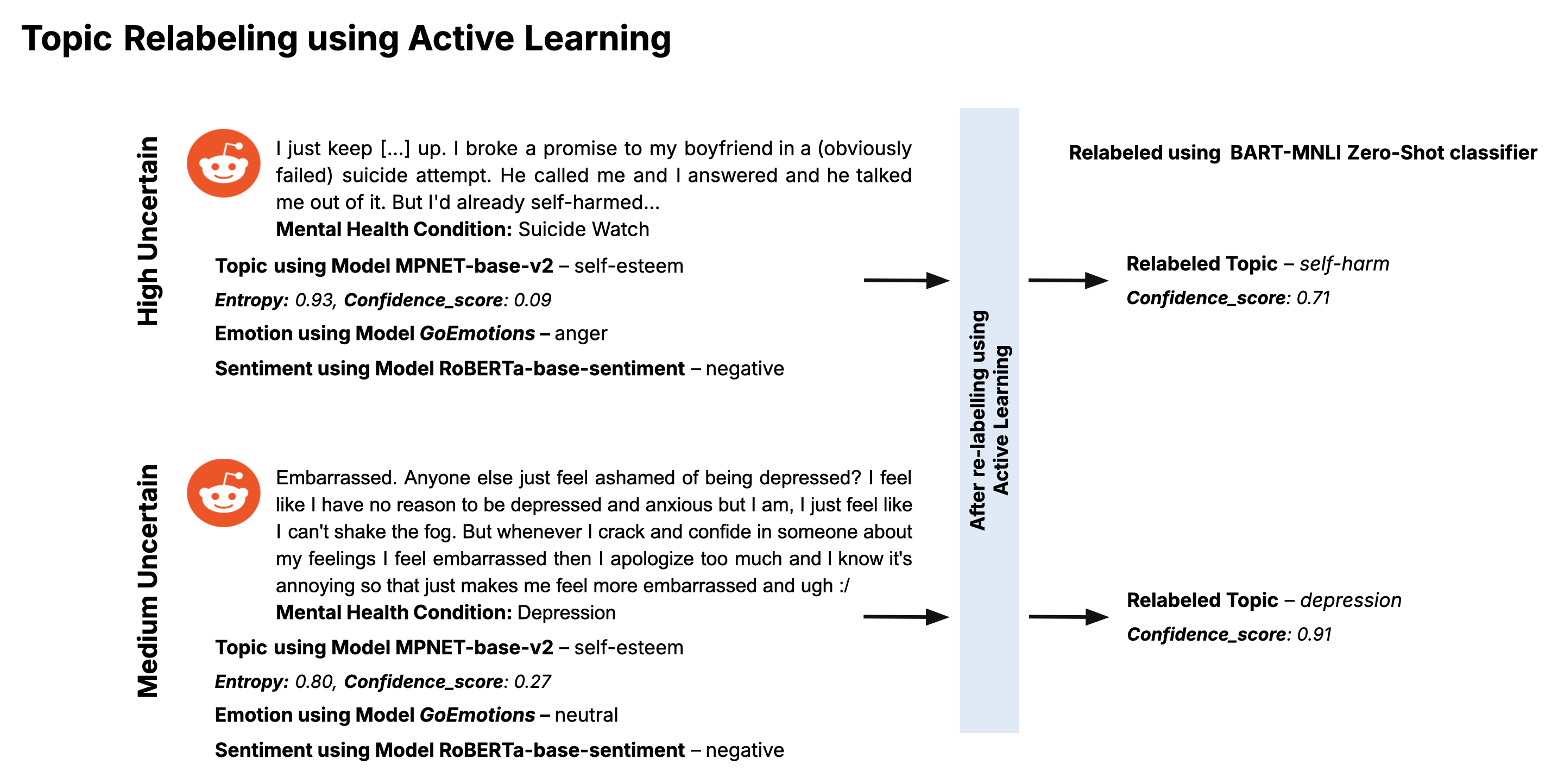}
    \caption{Topic relabeling using Medium and High Uncertainty sampling}
    \label{fig:topic_relabel_hm}
\end{subfigure}
\caption{\textbf{Examples of emotion and topic label corrections using single-stage Medium and High Uncertainty sampling.}}
\label{fig:relabel_hm_combined}
\end{figure}

\begin{figure}[H]
\centering
\begin{subfigure}[t]{\textwidth}
    \centering
    \includegraphics[width=\textwidth]{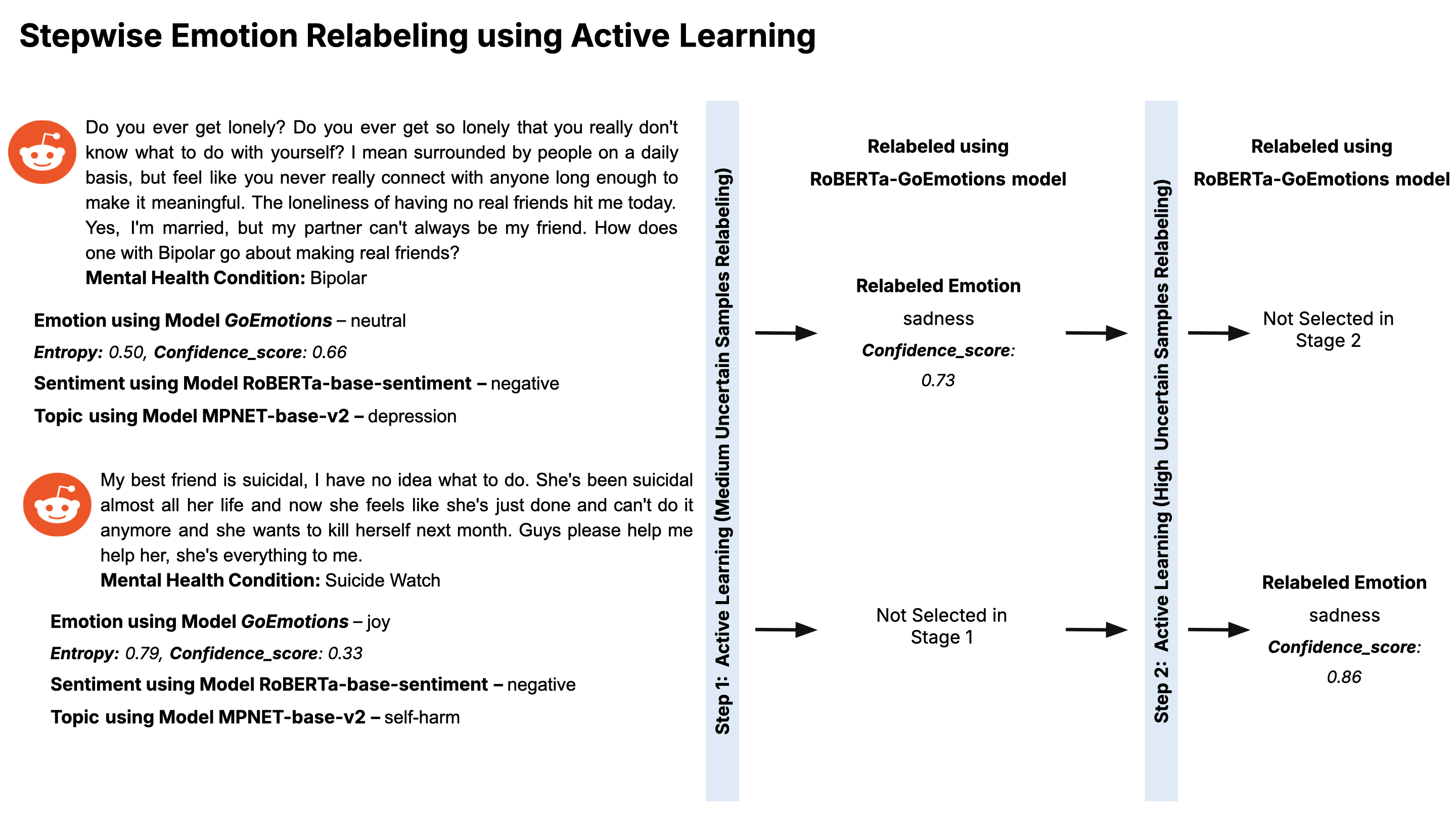}
    \caption{Emotion relabeling using Active Learning-based Stepwise Relabeling}
    \label{fig:emo_relabel_step}
\end{subfigure}

\vspace{0.5cm}

\begin{subfigure}[t]{\textwidth}
    \centering
    \includegraphics[width=\textwidth]{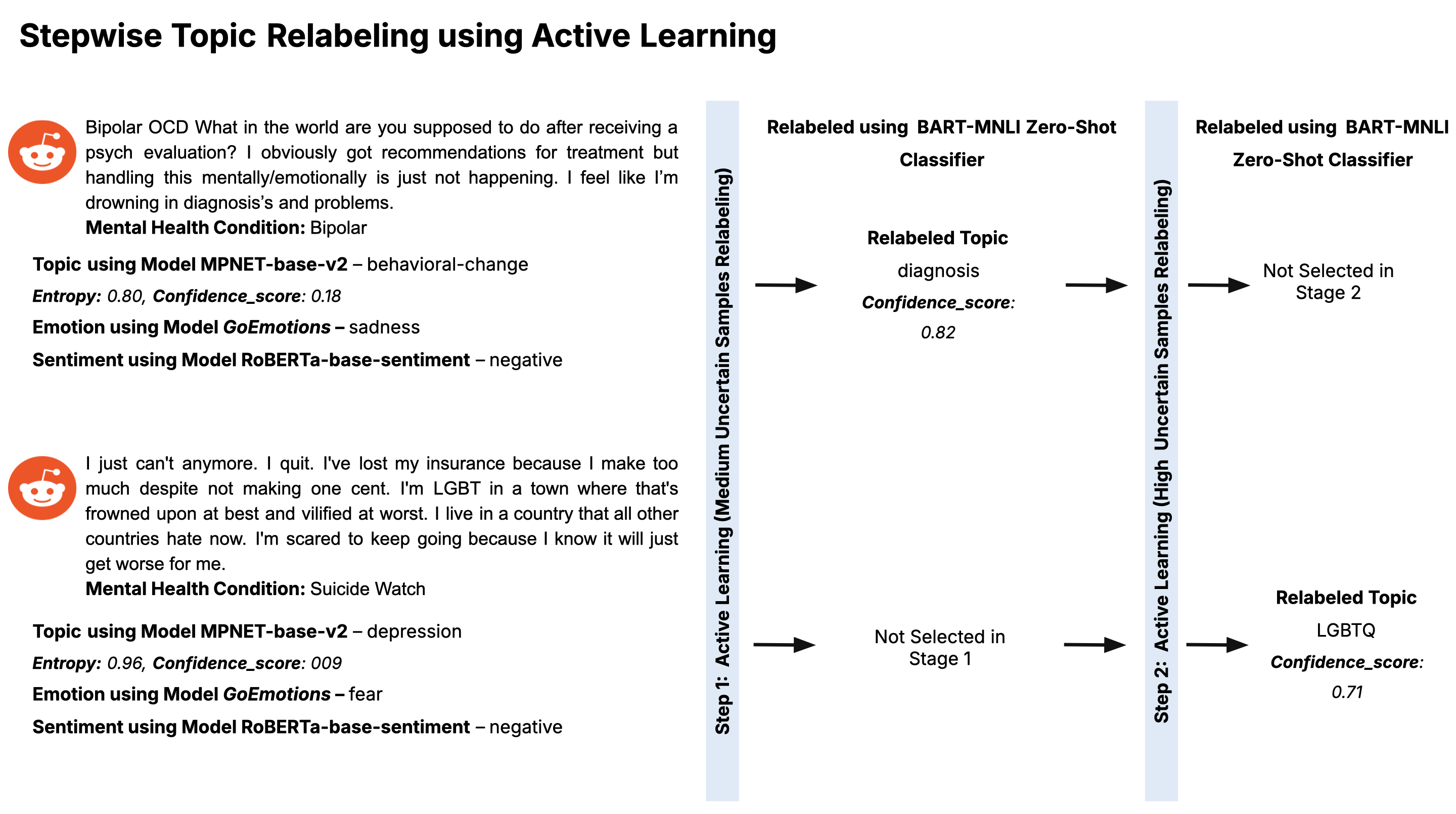}
    \caption{Topic relabeling using Active Learning-based Stepwise Relabeling}
    \label{fig:topic_relabel_step}
\end{subfigure}
\caption{\textbf{Examples of emotion and topic label corrections using Active Learning-based Stepwise Relabeling, with medium-uncertainty samples relabeled in Stage 1 and high-uncertainty samples relabeled in Stage 2.}}
\label{fig:relabel_step_combined}
\end{figure}

Figures~\ref{fig:relabel_hm_combined} and~\ref{fig:relabel_step_combined} show examples of emotion and topic labels corrected by each Active Learning strategy. In these examples, the original label generated by a pre-trained model is incorrect and does not match the post content. For example, in Figure~\ref{fig:relabel_hm_combined}, a \textit{Bipolar} post describing agoraphobia was labeled \textit{Neutral} and is corrected to \textit{Fear} (confidence: 0.74). A \textit{Suicide Watch} post describing a suicide attempt was given the topic \textit{Self-esteem} and is corrected to \textit{Self-harm} (confidence: 0.71). Figure~\ref{fig:relabel_step_combined} also shows how samples are selected for relabeling at different uncertainty levels in the two stages. For example, a post written by an LGBT user facing discrimination has an entropy of 0.96, outside the medium-uncertainty range of Stage 1, and is corrected from \textit{Depression} to \textit{LGBTQ} (confidence: 0.71) in Stage 2.

\subsection{Error Analysis}
\begin{figure}[H]
\centering
\includegraphics[width=\textwidth]{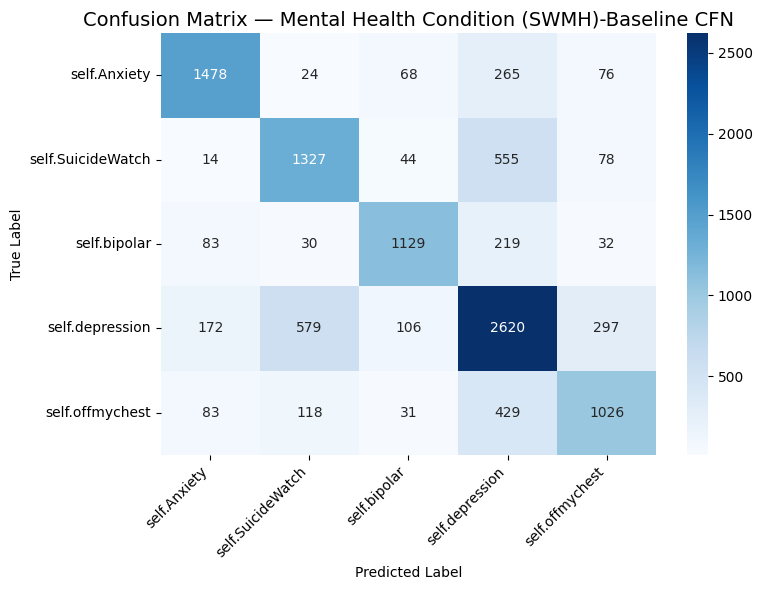}
\caption{\textbf{Confusion matrix for Mental Health Condition classification using the baseline CFN (MentalRoBERTa).}}
\label{fig:confmatrix_baseline}
\end{figure}

\begin{figure}[H]
\centering
\includegraphics[width=\textwidth]{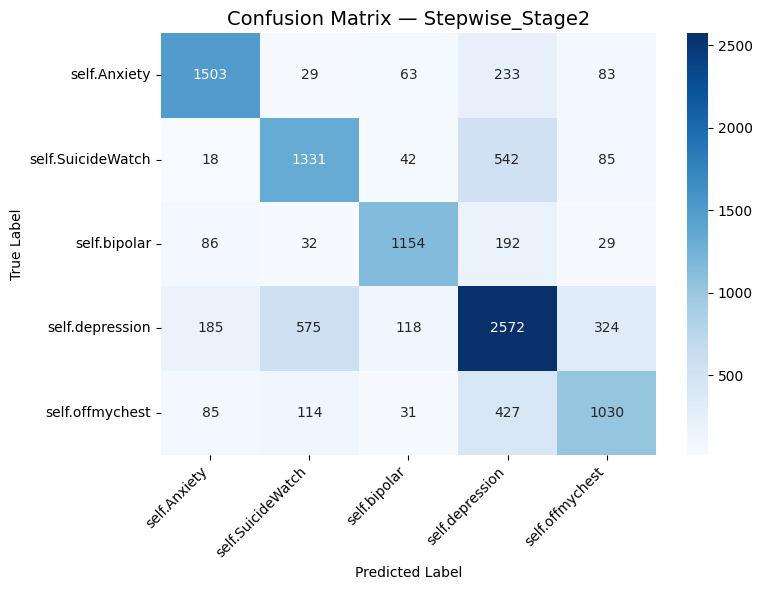}
\caption{\textbf{Confusion matrix for Mental Health Condition classification using the CFN (MentalRoBERTa) after Stepwise Relabeling Stage 2.}}
\label{fig:confmatrix_stepwise}
\end{figure}

Figures~\ref{fig:confmatrix_baseline} and~\ref{fig:confmatrix_stepwise} show the confusion matrices for the baseline CFN and Stepwise Relabeling Stage 2 on Mental Health Condition classification using MentalRoBERTa. The most frequent misclassification in both models is between \textit{Depression} and \textit{Suicide Watch}, with more than 540 posts misclassified in each direction. \textit{Off My Chest}, a general venting subreddit not tied to a specific condition, is the hardest class, with only 1,026 of its 1,687 posts classified correctly by the baseline and 1,030 by Stage 2, and over 420 predicted as \textit{Depression}. \textit{Anxiety} and \textit{Bipolar} are classified most accurately in both models. \textit{Depression}, the largest class, is the most common wrong prediction for every other class. Stage 2 misclassifies fewer posts as \textit{Depression} (1,394 vs.\ 1,468). In particular, fewer \textit{Anxiety} and \textit{Bipolar} posts are wrongly predicted as \textit{Depression}, so more of them are classified correctly. \textit{Suicide Watch} and \textit{Off My Chest} also improve slightly, while \textit{Depression} loses 48 correct predictions. Overall, Stage 2 improves four of the five classes, but confusion between \textit{Depression} and \textit{Suicide Watch} remains the main source of error.

\section{Limitations}
This work has a few limitations that point to useful directions for improvement. Even with the proposed Active Learning-based Stepwise Relabeling approach, Topic accuracy reaches only 0.6097, which reflects how challenging fine-grained topic classification becomes with 31 classes and an uneven distribution among them. The emotion, sentiment, and topic labels are generated automatically by pre-trained models rather than by human annotators. Relabeling refines only a subset of the emotion and topic labels, using pre-trained models, and leaves the sentiment labels unchanged, so some residual noise is likely still present. Since SWMH~\cite{swmh} draws from five Reddit mental health subreddits, the results reflect the language and reporting patterns of this specific online community, and testing the framework on other platforms or clinical settings would help confirm how well these findings generalize.

\section{Conclusion}
In this work, the three-task Core Fusion Network (CFN) is extended with Topic Classification as a fourth task, with topic labels derived from clinically grounded therapist-client interactions in the CounselChat~\cite{bertagnolli2020counsel} dataset. To address noise in the automatically generated emotion and topic labels, an Active Learning-based Stepwise Relabeling approach is proposed that progressively refines uncertain emotion and topic labels in two stages, first targeting medium-uncertainty samples and then high-uncertainty samples in Stage 2, while preserving previously learned task representations through fine-tuning rather than training from scratch. Extensive experiments on the large-scale SWMH~\cite{swmh} dataset of 54,412 Reddit posts, using both MentalBERT and MentalRoBERTa as encoders, show that the proposed approach improves all four tasks over the baseline CFN for both encoders. With MentalRoBERTa, accuracy improves by 0.09\% on Mental Health Condition, 0.51\% on Emotion, 0.35\% on Sentiment, and 1.57\% on Topic. The fine-tuned CFN also outperforms zero-shot and few-shot prompting of MentalLLaMA-7B on Mental Health Condition, Emotion, and Sentiment classification. The ablation study shows that only the stepwise approach improves all four tasks, and that Random Sampling has the lowest Mental Health Condition, Emotion, and Topic accuracy despite relabeling the same number of samples, confirming that entropy-based sample selection drives the gains. Qualitative examples further illustrate how relabeling corrects emotion and topic labels that do not match the post content. Overall, the results show that topic-aware multi-task learning combined with a stepwise relabeling approach produces more consistent improvements on noisy social media data than single-stage approaches.

Future work will explore relabeling a larger subset of the training data in each stage. Rather than accepting a label from a single pre-trained model above a fixed confidence threshold, ensemble relabeling models could change a label only when they agree, reducing label noise while limiting incorrect relabels. The most uncertain samples could also be verified by clinicians, moving the approach toward a human-in-the-loop Active Learning setting. Since a single post may express multiple emotions or address multiple topics, future work will treat the emotion and topic tasks as multi-label classification problems. In addition, the number of relabeling stages could be determined adaptively based on validation performance. Finally, evaluating the framework on other mental health datasets and in clinical settings would help assess its generalizability beyond Reddit-based social media text.


\bibliography{sn-bibliography}

\end{document}